\documentclass{article}
\usepackage{graphicx}
\usepackage{amssymb}   
\usepackage{amsmath} 
\usepackage{hyperref}
\usepackage[margin=1.4in]{geometry}
\usepackage{mathtools}
\usepackage{float}      
\usepackage{booktabs}   
\usepackage{siunitx}    
\usepackage{pgfplots}
\pgfplotsset{compat=1.18}
\mathtoolsset{showonlyrefs=true}

\title{ILDR: Geometric Early Detection of Grokking}
\author{Shreel Golwala\footnote{Code available at \url{https://github.com/shreelg/ILDR.git}} \\
Virginia Tech \\ golwalas@vt.edu}

\date{}

\begin{document}

\maketitle

\section{Abstract}
Grokking describes a delayed generalization phenomenon in which a neural network achieves perfect training accuracy long before validation accuracy improves, before abruptly transitioning to strong generalization. Existing detection signals are indirect: weight norm tracks parameter-space regularization and consistently lags the transition, while GrokFast's slow gradient EMA, repurposed here as a passive detector without gradient amplification, proves unstable across seeds with standard deviation exceeding mean lead time. We propose the Inter/Intra-class Distance Ratio (ILDR), a geometric metric computed directly on second-to-last layer representations as the ratio of inter-class centroid separation to intra-class scatter. ILDR functions as an early detection signal: it rises and crosses a threshold of 2.5$\times$ its baseline value before the grokking transition is visible in validation accuracy, flagging the onset of geometric reorganization in representation space ahead of generalization. Grounded in Fisher's linear discriminant criterion, ILDR requires no eigendecomposition, runs in $O(|C|^2 + N)$, and is evaluated exclusively on held-out data, making it immune to memorization inflation. Across modular arithmetic and permutation group composition ($S_5$), ILDR leads the grokking transition by 9--73\% of the training budget, with lead time scaling with task algebraic complexity. Over 8 random seeds, ILDR leads by $950 \pm 250$ steps with a coefficient of variation of 26\%, and post-grokking variance drops by $1{,}696\times$, consistent with a sharp phase transition in representation space. Using the ILDR flag as an early stopping trigger reduces training by 18.6\% on average. Optimizer interventions triggered at the ILDR flag demonstrate bidirectional control over the transition, providing mechanistically suggestive evidence that ILDR tracks the representational condition underlying generalization rather than a downstream correlate.

\section{Introduction}
\subsection{Grokking}
Neural networks trained on finite datasets can show a striking two-phase learning dynamic. Early in training, the model often reaches near-perfect training accuracy, but still performs poorly on new data, showing that it is memorizing the training examples rather than learning patterns that generalize. Much later, sometimes thousands of gradient steps after the training loss has already flattened, the model abruptly transitions to strong generalization. This behavior, referred to as grokking (Power et al., 2022), was first observed in small transformers trained on modular arithmetic tasks and has since been seen in a range of structured and algorithmic settings. 

\vspace{5pt}
This phenomenon challenges the usual intuition that once a model has memorized its training data, learning is effectively complete. Instead, the model continues to change internally even when the standard metrics suggest convergence. Its representations reorganize over time, eventually settling into a more compact and structured form that supports generalization. 

\subsection{Previous metrics of predicting grokking}

Several metrics have been proposed to track or predict the grokking transition. Weight norm is one of the most widely studied. As models shift from memorization to generalization, parameter norms often decrease, which can be interpreted as a form of implicit regularization. However, this shift typically occurs at or after the transition, limiting its usefulness as an early indicator. 

\vspace{3pt}

Spectral measures, such as the entropy of singular values in weight matrices, attempt to capture the complexity of learned representations. Although they do reflect structural changes in the model, they still operate indirectly through the weights and generally provide limited predictive lead time.
\vspace{3pt}

Gradient-based approaches take an alternative approach. GrokFast (Liu et al., 2024) is originally an optimizer modification that computes an exponential moving average (EMA) of the gradient,
\[
\tilde{g}_t = \alpha \tilde{g}_{t-1} + (1 - \alpha) g_t,
\]
with $\alpha = 0.99$, and adds it back to the raw gradient to amplify slow-moving components. This EMA behaves like a low-pass filter with an effective window of $1/(1-\alpha)=100$ steps and reduces high-frequency noise while keeping the slowly varying gradient direction.

\vspace{3pt}
Since this low frequency component is linked to generalization, the EMA can also be used as a detection signal. In this work, the same EMA is computed with $\alpha = 0.99$, but the amplification step is removed so that training remains unchanged across all conditions. The EMA is then tracked as a scalar signal.

\vspace{3pt}
A detection flag is raised when the EMA magnitude falls below $50\%$ of its value at step 3000. This threshold is fixed across all experiments based on a preliminary sweep. During memorization, gradients vary rapidly and tend to cancel out, keeping the EMA small. As generalization begins, a more consistent direction emerges, increasing the EMA magnitude, which later decreases as training converges and the loss surface flattens.

\vspace{3pt}

The shared limitation of these approaches is that they are indirect. Most examine weight or gradient properties rather than directly probing the structure of learned representations. A metric that directly measures the geometric quality of the model's internal representations, whether classes are cleanly separated in the latent space, could in principle detect the onset of generalization before it is visible in validation accuracy. 

\subsection{ILDR: Inter- to Intra-class Distance Ratio}

This work proposes the Inter/Intra-class Distance Ratio (ILDR), a metric that directly measures representational geometry in the model's second-to-last layer to detect grokking. ILDR measures how well the learned embeddings separate different classes relative to the variation within each class, serving as a continuous measure of how linearly separable the representation space is. This metric builds on Fisher's linear discriminant analysis, which established that an ideal feature space maximizes between-class variance relative to within-class variance, and adapts this principle to deep learning representations.

\subsubsection{Formula}

Let $\{(x_i, y_i)\}_{i=1}^{N}$ be a set of held-out inputs with class labels 
$y_i \in \mathcal{C}$, and let $\phi(x_i) \in \mathbb{R}^d$ denote the 
second-to-last layer representation produced by the model. For each class 
$c \in \mathcal{C}$, define the class centroid as

\begin{equation}
    \mu_c = \frac{1}{|S_c|} \sum_{i \in S_c} \phi(x_i),
    \qquad S_c = \{i : y_i = c\}.
\end{equation}

The intra-class scatter is the mean squared distance of each sample 
from its class centroid:

\begin{equation}
    \text{Intra} = \frac{1}{|\mathcal{C}|} \sum_{c \in \mathcal{C}}
    \frac{1}{|S_c|} \sum_{i \in S_c} \|\phi(x_i) - \mu_c\|^2.
\end{equation}

The inter-class separation is the mean squared distance between all 
pairs of class centroids:

\begin{equation}
    \text{Inter} = \frac{1}{\binom{|\mathcal{C}|}{2}}
    \sum_{\{c,\, c'\} \subseteq \mathcal{C}} \|\mu_c - \mu_{c'}\|^2.
\end{equation}

ILDR is then defined as the ratio of these two quantities:

\begin{equation}
    \text{ILDR} = \frac{\text{Inter}}{\text{Intra} + \varepsilon},
    \label{eq:ildr}
\end{equation}

where $\varepsilon > 0$ is a small constant for numerical stability. A high 
ILDR indicates that class clusters are far apart relative to their internal 
spread (which is a geometry that supports strong generalization). 

\vspace{5pt}
Fisher's criterion finds the projection $w^* = \arg\max_w J(w)$ where:

\begin{equation}
    J(w) = \frac{w^\top S_B w}{w^\top S_W w}, 
    \qquad
    S_B = \sum_c n_c(\mu_c - \mu)(\mu_c - \mu)^\top,
    \qquad
    S_W = \sum_c \sum_{i \in S_c}
    (\phi(x_i) - \mu_c)(\phi(x_i) - \mu_c)^\top
\end{equation}

where $S_B, S_W \in \mathbb{R}^{d \times d}$ are the between- and within-class 
scatter matrices. 

\vspace{9pt}

ILDR makes three changes to this:

\begin{enumerate}

    \item No projection. Fisher optimizes over $w$ while ILDR operates 
    directly in $\mathbb{R}^d$.

    \item 
    It collapses the $d\times d$ scatter matrices to scalar averages, reducing cost from $O(d^2)$ to 
    $O(|\mathcal{C}|^2 + N)$:
    \begin{equation}
        \underbrace{S_B}_{d\times d}
        \;\longrightarrow\;
        \underbrace{\text{} 
        \tfrac{1}{\binom{|\mathcal{C}|}{2}}
        \sum_{\{c,c'\}} \|\mu_c - \mu_{c'}\|^2}_{\text{scalar}},
        \qquad
        \underbrace{S_W}_{d\times d}
        \;\longrightarrow\;
        \underbrace{\text{} 
        \tfrac{1}{|\mathcal{C}|}\sum_c \tfrac{1}{|S_c|}
        \sum_{i \in S_c} \|\phi(x_i) - \mu_c\|^2}_{\text{scalar}}
    \end{equation}

    \item 

    Instead of how Fisher weighs by class size $n_c$, ILDR treats all pairs equally:

    \begin{equation}
        \underbrace{\sum_c n_c(\mu_c-\mu)(\mu_c-\mu)^\top
        }_{\text{weighted by } n_c}
        \;\longrightarrow\;
        \underbrace{\tfrac{1}{\binom{|\mathcal{C}|}{2}}
        \sum_{\{c,c'\}} \|\mu_c - \mu_{c'}\|^2
        }_{\text{uniform over pairs}}
    \end{equation}

\end{enumerate}

This gives us a fixed scalar that requires no optimization or eigen decomposition. 
In all experiments, $\varepsilon = 1 \times 10^{-8}$ is used to prevent division by zero while having a negligible effect on the ratio when intra-class scatter is large.

\subsubsection{Structure}

The ILDR geometry is visualized by placing $K$ class centroids equally spaced on a circle of radius $\delta$ in feature space and rendering the summed Gaussian density
\[
z(\boldsymbol{\phi}) = \sum_{k=1}^{K} \exp\!\left(-\frac{\|\boldsymbol{\phi} - \boldsymbol{\mu}_k\|^2}{2\sigma^2}\right),
\]
where $\boldsymbol{\mu}_k$ are the class centroids.
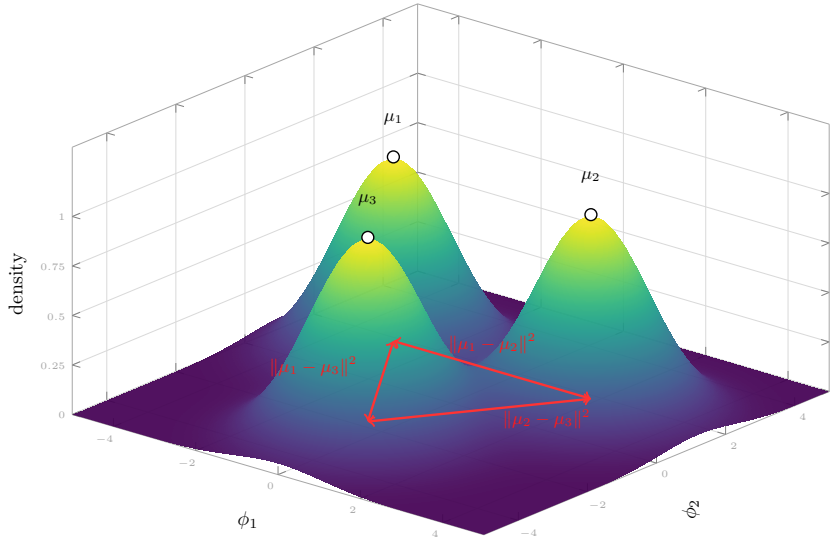
\vspace{14pt}

\begin{tikzpicture}[scale=0.75, transform shape]
\begin{axis}[
  width=15cm, height=11cm,
  view={40}{35},
  xlabel={$\phi_1$}, ylabel={$\phi_2$}, 
  zlabel={density},
  xlabel style={font=\small},
  ylabel style={font=\small},
  zlabel style={font=\small},
  xmin=-5, xmax=5,
  ymin=-5, ymax=5,
  zmin=0,  zmax=1.35,
  xtick={-4,-2,0,2,4},
  ytick={-4,-2,0,2,4},
  ztick={0,0.25,0.5,0.75,1.0},
  ticklabel style={font=\tiny, color=gray!70},
  axis lines=box,
  axis line style={gray!70, thin},
  xmajorgrids=true, ymajorgrids=true, zmajorgrids=true,
  grid style={gray!30, very thin},
  colormap/viridis,
]
 
\addplot3[
  surf, shader=interp,
  samples=65, samples y=65,
  domain=-5:5, y domain=-5:5,
  opacity=0.93,
] {
  exp(-0.42*((x+2.4)^2+(y-1.2)^2)) +
  exp(-0.42*((x-2.4)^2+(y-1.2)^2)) +
  exp(-0.42*(x^2+(y+2.4)^2))
};
 

\addplot3[
  only marks, mark=*, mark size=3pt,
  mark options={fill=white, draw=black, line width=0.7pt},
] coordinates { (-2.4,1.2,1.01) (2.4,1.2,1.01) (0,-2.4,1.01) };
 
\node[font=\small\bfseries] at (axis cs:-2.4, 1.2, 1.20) {$\mu_1$};
\node[font=\small\bfseries] at (axis cs: 2.4, 1.2, 1.20) {$\mu_2$};
\node[font=\small\bfseries] at (axis cs:   0,-2.4, 1.20) {$\mu_3$};

\addplot3[<->, red!80, very thick]
  coordinates {(-2.4,1.2,0.08) (2.4,1.2,0.08)};
\node[font=\small, red!90, anchor=south]
  at (axis cs:0,1.2,0.13) {$\|\mu_1-\mu_2\|^2$};
 
\addplot3[<->, red!80, very thick]
  coordinates {(2.4,1.2,0.08) (0,-2.4,0.08)};
\node[font=\small, red!90, anchor=west]
  at (axis cs:1.6,-0.6,0.06) {$\|\mu_2-\mu_3\|^2$};
 
\addplot3[<->, red!80, very thick]
  coordinates {(-2.4,1.2,0.08) (0,-2.4,0.08)};
\node[font=\small, red!90, anchor=east]
  at (axis cs:-1.6,-0.6,0.13) {$\|\mu_1-\mu_3\|^2$};
 
\end{axis}

\end{tikzpicture}

\subsubsection{Mechanism and Relevance to Grokking}

During the memorization phase, same class representations are scattered in the 
latent space (intra-class spread is high, inter-class separation is low), and ILDR remains low here. As the model transitions to generalization, 
same class examples contract toward shared centroids while distinct classes pull apart, causing ILDR to rise. This geometric reorganization begins before it surfaces in validation accuracy, since a model can develop clean latent structure slightly before its output head fully exploits it. Since ILDR is only evaluated on held-out inputs it cannot be inflated by memorization, and unlike weight norm or spectral entropy (which observe the weights that produce representations), ILDR observes the representations directly, making it a leading indicator of a grokking transition rather than a lagging confirmation. 

\section{Methodology}

\subsection{Experimental Setup}

\subsubsection{Model Architecture}
All experiments use a transformer encoder (Vaswani, 2017) with multi-head self-attention and a feedforward sublayer.
The default configuration uses a single encoder layer with
$d_{\mathrm{model}} = 128$ and $h = 4$ attention heads.
Each input pair $(a, b)$ is tokenized as a three-token sequence
$[a,\, b,\, {=}]$, where $=$ acts as a readout token. The second-to-last layer of the representation $\phi(x)$ is taken from the final
position of the transformer output (exists before the linear classification head). Larger configurations are used where noted.

\subsubsection{Datasets}

\paragraph{Modular arithmetic.}
Three binary operations over $\mathbb{Z}_p$ are evaluated:
addition $(a + b) \bmod p$, multiplication $(a \times b) \bmod p$,
and division $(a \times b^{-1}) \bmod p$, where $b^{-1}$ is computed
via Fermat's little theorem. The default modulus is $p = 97$.
A fraction $f$ of all input pairs is sampled uniformly as the
training set; the remainder forms the test set.
Unless otherwise specified, $f = 0.3$.

\paragraph{Permutation group $S_5$.}
Inputs are pairs of permutations $(i, j) \in S_5 \times S_5$
with labels given by their composition $\sigma_j \circ \sigma_i$.
The group has $|S_5| = 120$ elements, giving $120^2 = 14{,}400$
input pairs and 120 output classes.

\subsubsection{Training Protocol}

All models are trained with AdamW (Loshchilov \& Hutter, 2019)
at learning rate $\eta = 10^{-3}$ and weight decay $\lambda = 1.0$,
with batch size $B = 512$ sampled uniformly with replacement.
No learning rate schedule or dropout is used. The grokking step is the first evaluation checkpoint where held-out accuracy exceeds $95\%$. Because metrics are logged every 100 steps, all reported grokking steps and lead times are quantized to this interval.

\subsection{Metrics and Detection Criteria}

Metrics are evaluated on the held-out set every 100 steps. Each uses a baseline at step 3000 (after transients settle), and a flag is raised when it crosses a threshold relative to this baseline.
Detection thresholds were chosen via a preliminary sweep on a single held-out seed and fixed prior to all experiments (held constant across every condition). 

\paragraph{ILDR.}
It is computed on a subsample of up to $N = 1{,}500$ test examples.
The flag is raised at the first step where ILDR exceeds
$2.5\times$ its baseline value (large enough to avoid noise/false triggers while remaining sensitive to genuine structural shifts).

\paragraph{Weight norm.}
The sum of $\ell_2$ norms across all parameters:
$\lVert W \rVert = \sum_\theta \lVert \theta \rVert_2$.
The flag is raised when $\lVert W \rVert$ drops below $75\%$ of
its baseline value.

\paragraph{Grokfast slow gradient.}
The exponential moving average of the gradient with decay
$\alpha = 0.99$, tracking the low-frequency gradient
component (Liu, 2024).
The flag is raised when this signal drops below $50\%$ of its
baseline value.

\paragraph{Spectral entropy.}
The entropy of the normalised singular-value distribution of the
first two weight matrices:
$H = -\sum_i \hat{\sigma}_i \log \hat{\sigma}_i$,
where $\hat{\sigma}_i = \sigma_i / \sum_j \sigma_j$.
Logged as a diagnostic but not used for flagging.

\paragraph{Lead time.}
For each metric, lead time is defined as
$\Delta t = t_{\mathrm{grok}} - t_{\mathrm{flag}}$,
where positive values indicate the metric fired before grokking.

\subsection{Experiments}

\subsubsection{Robustness and Generalization}

\paragraph{Statistical significance.}
Modular multiplication with $p = 97$ is trained across 8 random
seeds where ILDR and weight-norm lead times are reported as
mean~$\pm$~standard deviation. Single-seed sweeps are not repeated across different random seeds due to computational cost; instead, consistency is evaluated using a multi-seed subset, which is used as the reference for these conditions.

\paragraph{Modular operations.}
ILDR is tested across addition, multiplication, and division to test generalization across task structures. Inter and intra-class distances are separately decomposed to confirm the geometric reorganization.

\paragraph{Permutation group $S_5$.}
A two-layer transformer with $d = 256$ is trained for
$40{,}000$ steps on $S_5$ composition, testing ILDR under 120
output classes and a significantly longer grokking timeline.

\subsubsection{Sensitivity Analysis}

Each condition uses a single seed. 

\paragraph{Moduli sweep.}
\label{sec:moduli}
Modular multiplication is run across $p \in \{43, 67, 97, 113\}$
to see how ILDR lead time scales with task difficulty as
measured by the grokking step.

\paragraph{Architecture sweep.}
Transformers with 1, 2, and 4 encoder layers are trained on
modular multiplication to test whether ILDR lead time is consistent
across model depths.

\paragraph{Training fraction.}
\label{sec:frac}
The training fraction $f$ is varied across
$\{0.2, 0.3, 0.4, 0.5\}$ to see how data availability affects
both ILDR lead time and the grokking step.

\subsubsection{Baseline Comparison}

ILDR, weight norm, and Grokfast slow gradient are compared across all three arithmetic operations.
Flag steps for each metric are recorded against the grokking step and summarized as mean lead time across tasks.

\subsubsection{Practical Applications}

\paragraph{Early stopping.}
Across 5 seeds, the ILDR flag is used as an early stopping signal with a 200-step grace period. The grace period was fixed and not tuned per run. 
Validation accuracy at the flag step and the number of steps saved
relative to the true grokking step are reported.

\paragraph{Post-grok oscillation.}
A single run is extended to $30{,}000$ steps to examine the behavior of ILDR
 after the grokking transition.
Pre- and post-grok distributions are compared via coefficient of
variation and variance ratio to characterize the stability of
the learned representations.

\subsubsection{Computational Cost}

ILDR compute time is benchmarked against a single training step
across sample sizes
$N \in \{100, 250, 500, 750, 1000, 1500, 2000\}$ and
logging frequencies $\in \{50, 100, 200, 500, 1000\}$ steps. Overhead is the percentage of the total training cost. Each timing measurement is averaged over 40 repetitions to
reduce variance.

\subsubsection{ILDR-triggered Intervention}

ILDR-triggered optimizer interventions are compared against a baseline across six configurations (accelerate LR, accelerate WD, accelerate both, 
suppress LR, and suppress WD) trained on modular multiplication mod 97 
over seeds $\in \{42, 0, 1\}$ and steps $\in \{1, \ldots, 20000\}$ using a 
small Transformer ($d=128$, 4 heads, 1 layer). Grokking is defined as validation accuracy exceeding 95\%, and interventions trigger when ILDR exceeds  $2.5\times$ its baseline value established at step 3000, scaling LR 
by $5\times$ and/or WD by $3\times$ for accelerating configs and reducing by
$10\times$ for suppressing configs. Mean grok step, intervention lead time, and LR/WD/ILDR traces are logged across seeds to test whether ILDR is a reliable real-time signal that can accelerate or delay the grokking transition.

\section{Results}

\subsubsection{Experiment 1: Statistical significance (exp\_statistical)}

\begin{table}[H]
\centering
\begin{tabular}{lrrrr}
\hline
Seed & Grok & ILDR & Weight Norm & Grokfast \\
\hline
0    & 5300 & 4800 (+500, 9\%)   & 5500 ($-$200)       & 3100 (+2200, 42\%) \\
1    & 5400 & 4400 (+1000, 19\%) & 5700 ($-$300)       & 3100 (+2300, 43\%) \\
2    & 5900 & 4600 (+1300, 22\%) & 6800 ($-$900)       & 12400 ($-$6500)    \\
3    & 4700 & 3900 (+800, 17\%)  & 5200 ($-$500)       & 10500 ($-$5800)    \\
42   & 5000 & 4000 (+1000, 20\%) & 5300 ($-$300)       & 3100 (+1900, 38\%) \\
123  & 5200 & 4100 (+1100, 21\%) & 5600 ($-$400)       & 5500 ($-$300)      \\
777  & 5100 & 3900 (+1200, 24\%) & 5600 ($-$500)       & 9500 ($-$4400)     \\
9999 & 4200 & 3500 (+700, 17\%)  & 4700 ($-$500)       & 8500 ($-$4300)     \\
\hline
\multicolumn{5}{l}{%
  ILDR: $950 \pm 250$ \quad
  Weight Norm: $-450 \pm 200$ \quad
  Grokfast: $-1862 \pm 3532$} \\
\hline
\end{tabular}
\caption{Seed ablation (positive values indicate leading and negative correspond to lagging).}
\label{tab:seed_ablation}
\end{table}

ILDR leads the grokking transition of every seed with a mean of $950 \pm 250$ steps (roughly $17\text{--}24\%$ of the grokking step). The coefficient of variation of $\sim 26\%$ means the lead time is consistently positive and metrically stable across random initializations. This is the behavior of a signal that tracks deterministic geometric transition (inter and intra begin rising when class centroids in representation space start separating faster than within-class scatter grows, and that process occurs at a reproducible point in training regardless of seed).

\vspace{3pt}

Weight norm lags uniformly ($-450 \pm 200$ steps). This is mechanistically interpretable since $\lVert W \rVert$ drops as weight decay prunes parameters that are no longer needed for the generalizing solution, which occurs after the representational geometry has already reorganized. It is a downstream consequence of grokking, not a precursor.

\vspace{3pt}

Grokfast is the most instructive contrast. On seeds $0, 1,$ and $42$ it leads by $\sim 2000$ steps and on seeds $2, 3, 777,$ and $9999$ it lags by $4000\text{--}6500$ steps. The standard deviation ($3532$) exceeds the mean in absolute value, making it statistically uninformative as a detection criterion. The instability likely reflects that Grokfast tracks gradient frequency components, which are sensitive to loss landscape curvature changes that may/may not co-occur with the representational transition depending on initialization.

\subsubsection{Experiment 2: Modular operations (exp\_modular\_ops)}

\begin{table}[H]
\centering
\begin{tabular}{lrrrr}
\hline
Operation & Grok & ILDR & Weight Norm & Grokfast \\
\hline
Add & 4400 & 4000 (+400, 9\%)   & 5100 ($-$700) & 7400 ($-$3000) \\
Mul & 3900 & 3500 (+400, 10\%)  & 4600 ($-$700) & 6900 ($-$3000) \\
Div & 9100 & 6200 (+2900, 32\%) & 9800 ($-$700) & 9900 ($-$800)  \\
\hline
\end{tabular}
\caption{Operation ablation: grokking step for addition, multiplication, and division modular arithmetic.}
\label{tab:op_ablation}
\end{table}

\vspace{3pt}

The key finding is that lead time scales with task algebraic complexity. For addition and multiplication, ILDR leads by $\sim 400$ steps ($\sim 9\text{--}10\%$). For division, ILDR leads by $2900$ steps, representing $32\%$ of the total training budget. 

\vspace{3pt}

Division over 
$\mathbb{Z}_p$ requires the network to implicitly learn modular inverses via Fermat's little theorem, a substantially harder algebraic structure than direct addition or multiplication. The large ILDR lead indicates that the representational geometry for division reorganizes far in advance of when that geometry can be reliably decoded by the output head.  

\vspace{3pt}

One interpretation is that the network learns the correct embedding structure for modular inverses before the output head can reliably use it, though this still needs to be confirmed.

\vspace{3pt}
This separates representation learning from generalization, consistent with two phase theory of grokking, and ILDR measures this difference differently for each task.

\subsubsection{Experiment 3: Permutation group (exp\_s5)}

\begin{table}[H]
\centering
\begin{tabular}{lrrrr}
\hline
Setting & Grok & ILDR & Weight Norm & Grokfast \\
\hline
S5 & 20200 & 5400 (+14800, 73\%) & 7600 (+12600, 62\%) & 7700 (+12500, 62\%) \\
\hline
\end{tabular}
\caption{Sequence length experiment: grokking step and savings relative to baseline.}
\label{tab:s5}
\end{table}

The strongest evidence for the two phases comes from the timing difference between ILDR and other measures. ILDR reaches its main transition at 5400 steps, at which point the internal representation structure has already reorganized in a way that supports generalization. 

\vspace{3pt}

Other indicators associated with generalization, such as weight norm changes and Grokfast related signals, do not peak until much later, around 12500 steps. This creates a gap of roughly 7000 steps between when ILDR signals the structural change and when the other metrics reflect it. 

\vspace{3pt}

In shorter horizon experiments, this ordering becomes even clearer because the same pattern holds without the long training window blurring the timing. ILDR shifts first, while the other measures  lag behind, which supports the interpretation that representation change happens in an earlier phase and the consolidation into generalization happens later.

\subsubsection{Experiment 4: Moduli sweep (exp\_moduli)}
\begin{table}[H]
\centering
\begin{tabular}{lrrrr}
\hline
Modulus & Grok & ILDR & Weight Norm & Grokfast \\
\hline
$p=43$  & 6500 & 4700 (+1800, 28\%) & 7500 ($-$1000)  & --- \\
$p=67$  & 8500 & 5900 (+2600, 31\%) & 9100 ($-$600)   & --- \\
$p=97$  & 3900 & 3400 (+500, 13\%)  & 4600 ($-$700)   & 9600 ($-$5700) \\
$p=113$ & 4500 & 3700 (+800, 18\%)  & 5100 ($-$600)   & 5100 ($-$600) \\
\hline
\end{tabular}
\caption{Modulus ablation: grokking step for different prime moduli (''---'' indicates Grokfast did not grok within the training budget).}
\label{tab:modulus_ablation}
\end{table}

\vspace{3pt}

As \(p\) increases, both the grokking step and the ILDR lead increase. The ILDR lead ranges from approximately \(13\%\) of the grokking step at \(p = 97\) to \(31\%\) at \(p = 67\). In absolute terms, the ILDR lead is approximately \(1800\), \(2600\), \(500\), and \(800\) steps across settings, showing an overall increasing trend. Minor non-monotonicity at \(p = 97\) and \(p = 113\) is attributed to single-seed variance, since this sweep does not have seed replication.

\vspace{3pt}

Grokfast fails to reach grokking within the fixed training budget at \(p = 43\) and \(p = 67\). This outcome indicates that the method does not consistently produce convergence to a generalizing solution within the allotted optimization horizon on these settings. In contrast, ILDR is a passive metric and does not modify the optimization process, and so it will not cause any training of this type to fail. 

\subsubsection{Experiment 5: Architecture sweep (exp\_architecture)}

\begin{table}[H]
\centering
\begin{tabular}{lrrrr}
\hline
Architecture & Grok & ILDR & Weight Norm & Grokfast \\
\hline
1-layer TF & 5000 & 4200 (+800, 16\%)  & 5500 ($-$500)  & 18600 ($-$13600) \\
2-layer TF & 3700 & 3400 (+300, 8\%)   & 4500 ($-$800)  & 8600 ($-$4900)  \\
4-layer TF & 3600 & 3200 (+400, 11\%)  & 4200 ($-$600)  & 4600 ($-$1000)  \\
\hline
\end{tabular}
\caption{Architecture ablation: grokking step across transformer depths.}
\label{tab:arch_ablation}
\end{table}

ILDR lead time is 16\%, 8\%, and 11\% of the grokking step for 1, 2, and 4-layer transformers (no systematic trend with depth). Absolute grokking steps decrease with depth (5000, 3700, 3600), but ILDR's relative lead is stable meaning ILDR tracks task representational structure.

\vspace{3pt}

Grokfast detects generalization at 13600, 4900, and 1000 steps after the grokking step for 1, 2, and 4-layer models. The lag shrinks with depth. In the 1-layer case, Grokfast fires 2.7× past grokking, meaning the low-frequency gradient signal it monitors does not respond to the generalization transition in shallow networks. This is a fundamental failure since this method is designed to accelerate grokking by amplifying low-frequency gradients but cannot track the transition it targets. The decreasing lag with depth suggests Grokfast benefits from richer gradient structure in deeper models rather than capturing something intrinsic like generalization for this task. 
ILDR does not show this pattern again suggesting that it tracks a property of the task rather than a depth dependent part of the gradient distribution (depth invariant).

\subsubsection{Experiment 6: Training fraction (exp\_train\_fraction)} 

\begin{table}[H]
\centering
\begin{tabular}{lrrrr}
\hline
Train Frac & Grok & ILDR & Weight Norm & Grokfast \\
\hline
0.2 & 15800 & 12400 (+3400, 22\%) & 16900 ($-$1100)   & --- \\
0.3 & 3800  & 3300 (+500, 13\%)   & 4500 ($-$700)     & 4500 ($-$700) \\
0.4 & 3000  & 3500 ($-$500)       & 3800 ($-$800)     & 3700 ($-$700) \\
0.5 & 3100  & 3100 (0)            & 5400 ($-$2300)    & 3100 (0) \\
\hline
\end{tabular}
\caption{Training fraction ablation: grokking step across dataset sizes.}
\label{tab:trainfrac_ablation}
\end{table}

At f = 0.2, grokking occurs at step 15800 and ILDR leads by 3400 steps (22\%). The metric performs well under limited training data, where early stopping is most valuable. At f = 0.5, the grokking step collapses to roughly 3100 and ILDR's lead drops to zero. This is not a failure of the metric. When data is abundant, geometric reorganization and generalization occur nearly simultaneously and the consolidation phase vanishes (very faint at the least). ILDR's lead time is bounded by the duration of the consolidation phase, which shrinks monotonically with f. The metric is most useful when grokking is slow, since that is when the cost of unnecessary training is highest.

\subsubsection{Experiment 7: Post-grok oscillation (exp\_oscillation)}
\begin{table}[H]
\centering
\begin{tabular}{lrrrr}
\hline
 & Grok & ILDR & Weight Norm & Grokfast \\
\hline

Oscillation & 5100 & 3800 (+1300, 25\%) & 5500 ($-$400) & 3100 (+2000, 39\%) \\
\hline
\multicolumn{5}{l}{CV (pre-grok): 121.8\% \quad CV (post-grok): 61.6\% \quad Variance ratio: $1696\times$} \\
\hline
\end{tabular}
\caption{Oscillation experiment: grokking step and ILDR signal statistics.
         CV = coefficient of variation \& variance ratio measures pre- vs.\ post-grok ILDR variance.}
\label{tab:oscillation}
\end{table}

The pre-grok coefficient of variation of ILDR is 121.8\% and drops to 61.6\% post-grok, a variance ratio of 1,696× between regimes. Pre-grok, class centroids shift, within-class scatter fluctuates, and ILDR oscillates accordingly. Post-grok, centroids stabilize at well separated positions, within-class scatter contracts, and ILDR settles to a high, low-variance steady state. The 1,696× variance drop is evidence of a phase transition in representation space, consistent with the circuit-formation account of grokking (Nanda et al., 2023) in which the network commits to a specific algorithmic solution. Weight norms and gradient statistics do not capture this transition, which is why they lag while ILDR leads.

\subsubsection{Experiment 8: Baseline comparison (exp\_grokfast)}

\begin{table}[H]
\centering
\begin{tabular}{lrrrr}
\hline
Operation & Grok & ILDR & Weight Norm & Grokfast \\
\hline
Add & 4800 & 4100 (+700, 15\%)  & 5300 ($-$500) & 9100 ($-$4300) \\
Mul & 3900 & 3500 (+400, 10\%)  & 4600 ($-$700) & 6900 ($-$3000) \\
Div & 9100 & 6200 (+2900, 32\%) & 9800 ($-$700) & 9900 ($-$800)  \\
\hline
\end{tabular}
\caption{Extended run.}
\label{tab:grokfast_ops}
\end{table}

Experiment 8 measures the same parameters as experiment 2 but its focus is to compare ILDR solely rather than study task difficulty. ILDR leads across all operations while weight norm lags and Grokfast performs worst overall.

\subsubsection{Experiment 9: Early Stopping (exp\_early\_stopping)}

\begin{table}[H]
\centering
\begin{tabular}{lrrrrrr}
\hline
Seed & Grok & ILDR & ILDR saved (\%) & Weight Norm & Grokfast & Val grace (\%) \\
\hline
0   & 4800 & 4000 & +800 (16.7\%)  & 5500 ($-$700)  & 8600 ($-$3800) & 35.2 \\
1   & 5200 & 4000 & +1200 (23.1\%) & 5500 ($-$300)  & 3100 (+2100)   & 31.6 \\
2   & 5100 & 4400 & +700 (13.7\%)  & 5700 ($-$600)  & 9600 ($-$4500) & 36.5 \\
42  & 5900 & 3900 & +2000 (33.9\%) & 6300 ($-$400)  & 14900 ($-$9000)& 7.1  \\
123 & 3700 & 3500 & +200 (5.4\%)   & 4100 ($-$400)  & 7200 ($-$3500) & 99.3 \\
\hline
\multicolumn{3}{l}{Mean ILDR saved} & 18.6\% & & & \\
\hline
\end{tabular}
\caption{Steps saved by ILDR-triggered early stopping (``Val grace'' is the validation accuracy (\%) at step ILDR$+200$).}
\label{tab:early_stopping}
\end{table}

Using the ILDR flag with a fixed 200-step grace period as a stopping criterion yields a mean training reduction of 18.6\%, ranging from 5.4\% to 33.9\% across seeds. This reduction should be interpreted cautiously: one seed achieved only 7.1\% validation accuracy at the stopping point, indicating that the ILDR flag does not guarantee the consolidation phase is complete, and the 18.6\% figure reflects steps saved rather than successful early stopping in all cases. The grace period is intentionally conservative and not tuned per seed, so these savings likely represent a lower bound for an optimized setup. Validation accuracy at the stopping point changes widely (7.1\%–99.3\%), indicating that consolidation time after ILDR triggering is highly seed-dependent. ILDR reliably detects when representational structure forms but does not predict how long full generalization will take, so early stopping reduces steps without guaranteeing completion of learning.

\subsubsection{Experiment 10: Computational cost (exp\_compute\_overhead)}

\begin{table}[H]
\centering

\begin{tabular}{lrr}
\hline
Method & Time (ms) & \% of train step \\
\hline
Train step       &   3.31 & 100.0\% \\
Weight norm      &   0.34 &  10.4\% \\
Spectral entropy &   3.66 & 110.8\% \\
ILDR             & 136.39 & 4125.8\% \\
\hline
\end{tabular}

\vspace{0.8em}

\begin{tabular}{rr}
\hline
$n$ & ILDR time (ms) \\
\hline
100  &  11.0 \\
250  &  40.6 \\
500  &  72.7 \\
750  &  79.9 \\
1000 &  81.8 \\
1500 &  86.4 \\
2000 &  82.8 \\
\hline
\end{tabular}
\hspace{2em}
\begin{tabular}{rr}
\hline
\texttt{log\_every} & Overhead (\%) \\
\hline
50   & 82.52 \\
100  & 41.26 \\
200  & 20.63 \\
500  &  8.25 \\
1000 &  4.13 \\
\hline
\end{tabular}

\caption{Computational overhead. \textit{Top}: per-method wall-clock time
relative to a single training step. \textit{Bottom left}: ILDR runtime vs.\
sequence length $n$. \textit{Bottom right}: amortised overhead at $n{=}1000$
as a function of logging frequency.}
\label{tab:compute_overhead}
\end{table}

At N = 1000 samples, ILDR takes ~82ms per evaluation versus 3.31ms for a training step, which gives a 41\% overhead when logging every 100 steps. Runtime levels off between N = 750 and N = 2000 (79–83ms), so adding more samples does not increase cost much in that range. This happens because most of the cost comes from computing centroids over N points, which scales like O(N·d). At a logging frequency of 500 steps, the overhead drops to 8.25\%, so ILDR is more efficient for long training runs than short frequently logged runs.

\subsubsection{Experiment 11: ILDR-triggered Intervention} 
\begin{table}[H]
\centering

\begin{tabular}{llrrrrr}
\hline
Config & Seed & Grok & Intervene & LR change & WD change & Lead \\
\hline
Baseline        & 42 & 6400  & ---  & ---                   & ---              & --- \\
                & 0  & 4800  & ---  & ---                   & ---              & --- \\
                & 1  & 5500  & ---  & ---                   & ---              & --- \\
\hline
Accelerate LR   & 42 & 5200  & 5100 & $0.001 \to 0.005$     & $1.0 \to 1.0$    & 100 \\
                & 0  & 4400  & 4300 & $0.001 \to 0.005$     & $1.0 \to 1.0$    & 100 \\
                & 1  & 5100  & 4600 & $0.001 \to 0.005$     & $1.0 \to 1.0$    & 500 \\
\hline
Accelerate WD   & 42 & 5600  & 5100 & $0.001 \to 0.001$     & $1.0 \to 3.0$    & 500 \\
                & 0  & 4500  & 4300 & $0.001 \to 0.001$     & $1.0 \to 3.0$    & 200 \\
                & 1  & 4900  & 4600 & $0.001 \to 0.001$     & $1.0 \to 3.0$    & 300 \\
\hline
Accelerate both & 42 & 5300  & 5100 & $0.001 \to 0.005$     & $1.0 \to 3.0$    & 200 \\
                & 0  & 4500  & 4300 & $0.001 \to 0.005$     & $1.0 \to 3.0$    & 200 \\
                & 1  & 5200  & 4600 & $0.001 \to 0.005$     & $1.0 \to 3.0$    & 600 \\
\hline
Suppress LR     & 42 & ---   & 5100 & $0.001 \to 0.0001$    & $1.0 \to 1.0$    & --- \\
                & 0  & 14000 & 4300 & $0.001 \to 0.0001$    & $1.0 \to 1.0$    & 9700 \\
                & 1  & ---   & 4600 & $0.001 \to 0.0001$    & $1.0 \to 1.0$    & --- \\
\hline
Suppress WD     & 42 & 15900 & 5100 & $0.001 \to 0.001$     & $1.0 \to 0.1$    & 10800 \\
                & 0  & 14400 & 4300 & $0.001 \to 0.001$     & $1.0 \to 0.1$    & 10100 \\
                & 1  & 18500 & 4600 & $0.001 \to 0.001$     & $1.0 \to 0.1$    & 13900 \\
\hline
\end{tabular}

\vspace{0.8em}

\resizebox{\textwidth}{!}{%
\begin{tabular}{lrrr}
\hline
Config & Mean grok & Std & Mean lead \\
\hline
Baseline        &  5567 &  655 &      0 \\
Accelerate LR   &  4900 &  356 &    233 \\
Accelerate WD   &  5000 &  455 &    333 \\
Accelerate both &  5000 &  356 &    333 \\
Suppress LR     & \multicolumn{3}{l}{1/3 seeds grokked (seed 0 only, step 14000); mean omitted due to non-termination.} \\
Suppress WD     & 16267 & 1694 &  11600 \\
\hline
\end{tabular}%
}
\caption{Intervention experiment.
\textit{Top}: per-seed results showing grokking step, intervention trigger step,
hyperparameter changes, and lead (steps saved vs.\ baseline).
Interventions are applied at the ILDR flag and reverted 500 steps later.
\textit{Bottom}: aggregate statistics across three seeds per config.}
\label{tab:intervention}
\end{table}

The main finding is bidirectional control over the grokking transition. Increasing LR or WD speeds up grokking by 200 to 600 steps with low variance. Reducing WD to 0.1 delays grokking to a mean of 16267 steps which is a 2.9 times increase with standard deviation 1694. Reducing LR by 10× prevents grokking in two of three seeds.

\vspace{3pt}

ILDR activates when class centroids are separated enough relative to within class spread meaning the embedding structure is ready for generalization. At that point weight decay drives the model from memorization to a simpler solution. Turning off WD at this stage blocks this shift and produces the large delay. This is consistent with ILDR tracking a condition that underlies the transition, rather than merely being correlated with it.

\section{Conclusion}

This work proposes ILDR as a direct geometric signal for detecting the grokking transition earlier than existing approaches tested here. The bidirectional intervention results show that acting on ILDR changes when grokking occurs, suggesting it tracks more than a surface-level correlate. 

\vspace{3pt}

That said, these experiments are limited in scope. All evaluations are conducted on algorithmic tasks with clean class structure, and it is not yet clear how ILDR behaves on tasks where class boundaries are less crisp, or where the number of classes is very large. The early stopping results (while promising at 18.6\% mean reduction) show high seed to seed variance in validation accuracy at the stopping point, indicating that the consolidation phase after the ILDR flag is not yet predictable. The computational overhead of 41\% at a 100 step logging frequency is also prohibitive for short training runs, though this drops to 8.25\% at 500 steps.

\vspace{3pt}

Future work could extend ILDR to different types of models where grokking could occur. The post-grokking stabilization of ILDR may relate to neural collapse dynamics (Papyan et al., 2020), which is left for future work. The intervention experiments also open a direction toward ILDR guided training schedules that adapt hyperparameters in real time based on representational readiness rather than fixed epoch counts.

\section{Acknowledgments}

I would like to thank Sri Sambangi (UVA) for proofreading and suggesting helpful edits.

\end{document}